\pdfoutput=1

\documentclass[11pt]{article}

\usepackage[preprint]{acl}

\usepackage{times}
\usepackage{latexsym}

\usepackage[T1]{fontenc}

\usepackage[utf8]{inputenc}

\usepackage{microtype}

\usepackage{inconsolata}

\usepackage{graphicx}

\usepackage{booktabs}
\usepackage{float}
\usepackage{tabularx}
\usepackage{makecell} 
\usepackage{subcaption}
\usepackage{multirow}
\usepackage{authblk}

\title{Enhancing BERTopic with Intermediate Layer Representations}

\begin{document}
\author[1]{Dominik Koterwa}
\author[1]{Maciej Świtała}
\affil[1]{Faculty of Economic Sciences, University of Warsaw}
\affil[ ]{\small\textbf{Correspondence:} \href{mailto:koterwadominik@icloud.com}{koterwadominik@icloud.com}}

\maketitle
\begin{abstract}
BERTopic is a topic modeling algorithm that leverages Transformer-based embeddings to create dense clusters, enabling the estimation of topic structures and the extraction of valuable insights from a corpus of documents. This approach allows users to efficiently process large-scale text data and gain meaningful insights into its structure. While BERTopic is a powerful tool, embedding preparation can vary, including extracting representations from intermediate model layers and applying transformations to these embeddings. In this study, we evaluate 18 different embedding representations and present findings based on experiments conducted on three diverse datasets. To assess the algorithm’s performance, we report topic coherence and topic diversity metrics across all experiments. Our results demonstrate that, for each dataset, it is possible to find an embedding configuration that performs better than the default setting of BERTopic. Additionally, we investigate the influence of stop words on different embedding configurations.

\end{abstract}

\section{Introduction}

Topic modeling (TM) is a popular tool in text mining which attempts to process multiple unlabeled documents and provide valuable insights about their interior, mainly by creating a representation of topics present in them. 
Throughout the years many topic modeling algorithms have been developed and applied in various fields, such as finance \cite{aziz2022tmfinance, ogunleye2023tmbanking, raju2022tmfinancialbureau}, social media \cite{falkenberg2022tmsocialmediaclimatechange, li2023tmchatgptsocialmedia, ramondt2022smblooddonation}, or medicine \cite{jeon2023tmpsychiatry, yao2018tmhinesemedicine, guizzardi2023tmboneregeneration}. Topic modeling has proven to be a useful tool to retrieve valuable insights about data from different modalities, for example images \cite{feng2010tmimages}, video \cite{hospedales2012tmvideo}, and audio \cite{gong2017tmaudio}. Ongoing advancements are being pursued in the field of topic modeling, with novel methodologies and insights emerging progressively over time. 
\newline
\indent 
One of the more recent advancements in topic modeling is BERTopic \cite{grootendorst2022bertopic}, a technique that leverages embeddings from Sentence Transformers \cite{reimers2019sentencebert} and clustering methods to generate coherent and interpretable topics. Sentence Transformers typically consist of multiple stacked encoder layers that process information sequentially. Each encoder layer generates intermediate representations known as \textit{hidden states}. Unlike traditional approaches such as Latent Dirichlet Allocation (LDA) \cite{blei2003latent} or Non-negative Matrix Factorization (NMF) \cite{paatero1994positive}, BERTopic incorporates contextualized word representations, allowing it to capture semantic relationships more effectively. This makes it particularly useful for analyzing large-scale text data across various domains. However, despite its growing popularity, the impact of different embedding representations for document encoding has not been yet studied. Most implementations rely on the default setting (Mean pooling with the last layer’s hidden states) without exploring alternative strategies that could potentially enhance quality.

\section{Related work}
This research is mainly inspired by studies examining the geometry and structure of contextualized word representations. Several works have analyzed the influence of embeddings from different layers of Transformer-based models \cite{vaswani2017attention}. \citet{ethayarajh2019contextual} investigated the contextualization of word representations and the variations in encoding the same words in different contexts. \citet{ma2019universal} evaluated embeddings extracted from various encoder layers on a range of downstream and probing tasks, demonstrating that upper-layer embeddings generally yield better performance. However, for certain probing tasks, such as tense classification as well as subject and object number classification, representations from middle layers achieve the best results. The study also examined the effect of pooling methods on model performance, showing significant differences in metric values between various pooling strategies.
\citet{jawahar2019bertstructureoflanguage} visualized how BERT embeddings shift across layers and conducted multiple probing experiments, concluding that different layers vary in their ability to encode syntactic and semantic information. Similarly, \citet{puccetti2021bertorganizeknowledge} applied logistic regression \cite{tibshirani1996lassoregression} to BERT activations and presented results on 68 sentence-level probing tasks. Their findings suggest that specific groups of hidden units are more relevant to particular linguistic properties and similar tasks. The influence of pooling strategies was also evaluated.
\citet{raganato2018analysismultilingual} explored Transformer representations in a multilingual setting, revealing variations in probing task performance across different layers for all languages examined.
\newline
\indent
Inspired by previous research, this study investigates whether embeddings extracted from different model layers significantly affect evaluation metrics in a topic modeling setting and whether alternative strategies lead to improved results. To explore these questions, we adopt the same architectural framework as in BERTopic paper.

\section{Data}
We have decided to use exactly the same datasets evaluated by \citet{grootendorst2022bertopic} to make the results comparable between two works.

\begin{table}[h]
\centering
\tiny 
\setlength{\tabcolsep}{4pt} 
\renewcommand{\arraystretch}{1.3} 
\begin{tabular}{c c c c c} 
\toprule
\makecell[c]{Dataset} & \makecell[c]{\# Documents} & \makecell[c]{Vocab \\ Size} & \makecell[c]{Avg \\ Words} & \makecell[c]{Most Common Words} \\
\midrule
20 Newsgroups & 16,309 & \makecell[c]{1,612} & \makecell[c]{48.2} & \makecell[c]{make, people, time \\ good, work, year \\ system, file, find \\ give} \\
\midrule
Trump Tweets & 44,252 & \makecell[c]{20,319} & \makecell[c]{15.86} & \makecell[c]{great, trump, thank \\ people, would, get \\ new, president, like \\ big} \\
\midrule
United Nations & 50,422 & \makecell[c]{29,157} & \makecell[c]{68.38} & \makecell[c]{united, security, nations \\ development, world, must \\ international, also, peace \\ countries} \\
\bottomrule
\end{tabular}
\caption{Descriptive statistics of datasets used in experiments.}
\label{tab:table1}
\end{table}

\indent
The BBC News dataset \cite{greene2006bbcnewsdataset} was replaced with transcriptions of United Nations (UN) General Debates from 2006 to 2015\footnote{\url{https://runestone.academy/runestone/books/published/httlads/_static/un-general-debates.csv}}. The UN debates corpus was previously used in the BERTopic article but specifically in the dynamic topic modeling setting. This substitution was made due to the limited size of the BBC News dataset, which contained only 2,225 observations. During evaluation, the BERTopic algorithm reduces the number of topics from an initially calculated value to a user-specified parameter. For this dataset, the evaluation tool consistently detected fewer topics than the user-defined parameter. As a result, the calculated results showed minimal variation in nearly all tested configurations.
\newline
\indent
Trump’s tweets\footnote{\url{https://www.thetrumparchive.com/faq}} from 2009 to 2021, totaling 44,252 observations (excluding retweets), serve as a source of short-text data. Since BERTopic has been applied to tweets from various domains, analyzing this dataset can provide valuable insights across multiple fields.

The 20 Newsgroups dataset\footnote{\url{https://github.com/MIND-Lab/OCTIS/tree/master/preprocessed_datasets/20NewsGroup}} consists of 16,309 news articles across 20 categories. This dataset complements the other two, allowing for evaluations across three distinct domains. More details on the structure of the datasets are provided in Table \ref{tab:table1}.

\begin{figure}
    \centering
    \begin{subfigure}{\columnwidth}  
        \centering
        \includegraphics[width=\columnwidth]{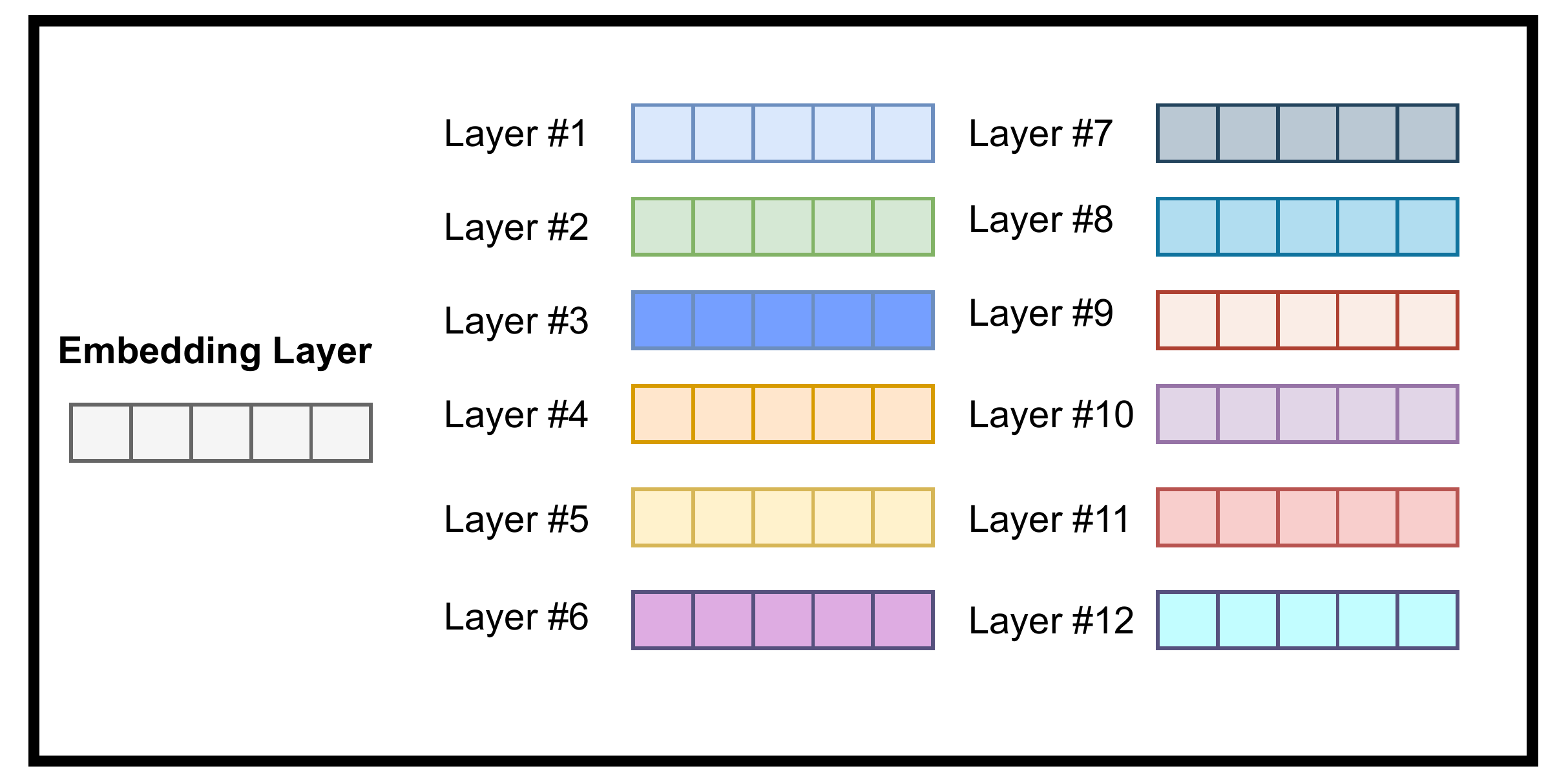}
        \caption{Possible to retrieve embeddings from the model.}
    \end{subfigure} 
    \vspace{1em}  
    \begin{subfigure}{\columnwidth}  
        \centering
        \includegraphics[width=\columnwidth]{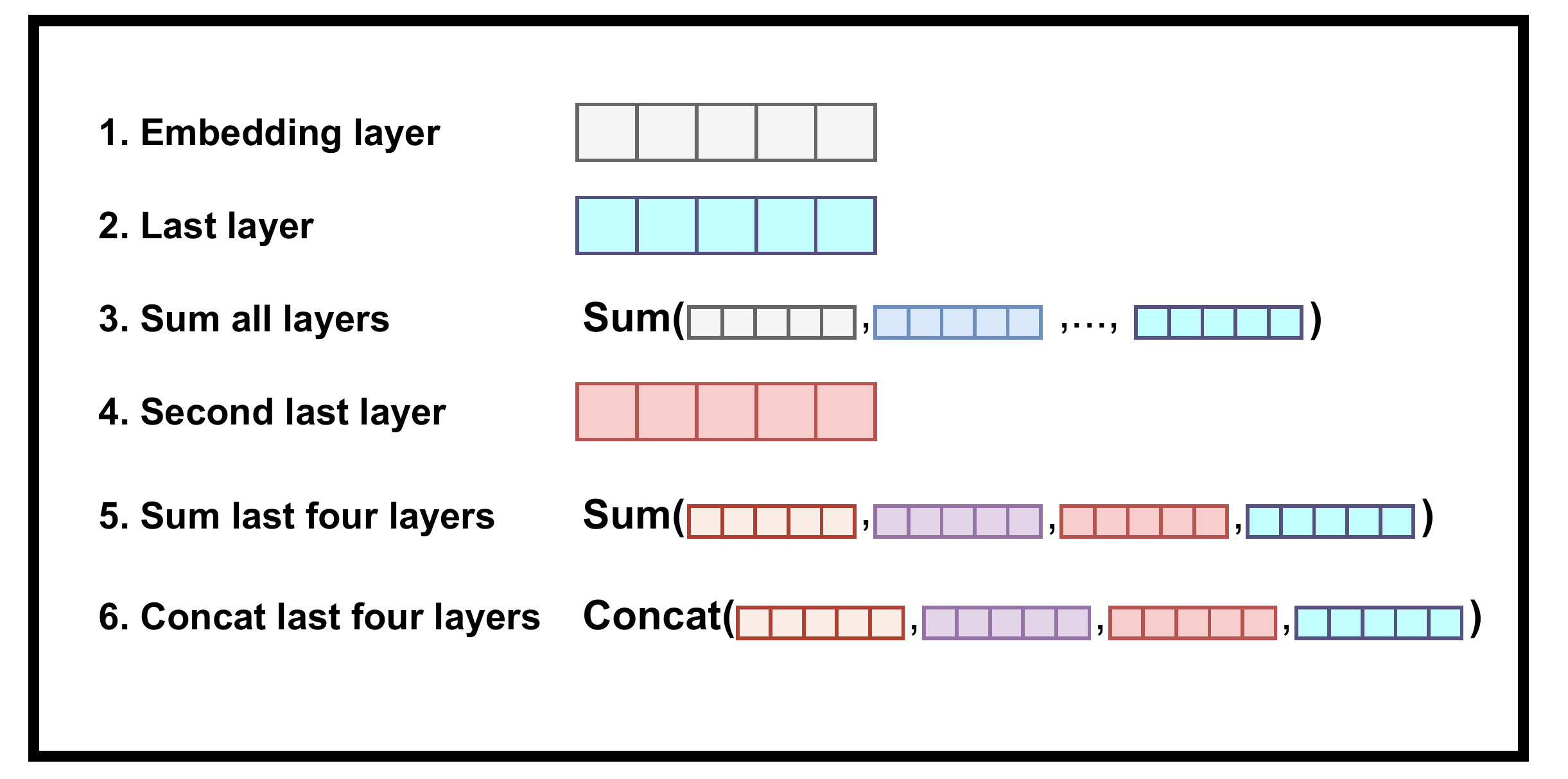}
        \caption{Configurations of embeddings used in experiments.}
    \end{subfigure}
    \caption{Visualization of embedding configurations used in our experiments, "Sum" refers to summing the embeddings, while "Concat" indicates concatenating the representations.}
    \label{fig:fig1}
\end{figure}

\section{Experiments}
Our embedding configurations closely follow those used by \citet{devlin2018bert}, with six distinct retrieval methods implemented.
Embedding retrieval is performed by extracting hidden states from specific encoders. We then apply functions such as summation or concatenation to construct final embeddings for evaluation. The following representations are used:
\begin{enumerate} \item Output from the last layer
\item Output from the embedding layer
\item Sum of outputs from all layers
\item Output from the second-to-last layer
\item Sum of outputs from the last four layers
\item Concatenation of outputs from the last four layers
\end{enumerate}

An illustration of the model’s possible outputs and the evaluated configurations is provided in Figure \ref{fig:fig1}. In addition, three pooling methods—Mean, Max, and CLS pooling—were incorporated into the experimental setup, resulting in a total of 18 configurations. To retrieve embeddings from various layers, we used the same model as \citet{grootendorst2022bertopic}, specifically the \textit{all-mpnet-base-v2} model from the Sentence Transformers library\footnote{\url{https://www.sbert.net/\#}}. The evaluation was conducted using the pipeline from BERTopic’s author\footnote{\url{https://github.com/MaartenGr/BERTopic_evaluation}}.
We assess the performance of the topic modeling algorithm using topic diversity and topic coherence, two widely used metrics in topic modeling research. Topic coherence is measured using normalized pointwise mutual information (NPMI) \cite{bouma2009normalized}, where the scale ranges from -1 to 1, with 1 indicating an ideal relationship. Topic diversity quantifies the percentage of unique words among the top words of all topics \cite{dieng2020topicdiversity}, with a diversity close to 0 indicating redundant topics and a diversity close to 1 indicating more varied topics. While overall quality is sometimes calculated as the product of topic coherence and diversity, we report these metrics separately in our analysis. It is important to note that commonly reported metrics in topic modeling may not fully capture algorithm quality. These metrics can provide useful indications, but a more thorough self-made manual assessment is needed to evaluate the results comprehensively.
Experiments were conducted for topic counts ranging from 10 to 50, with steps of 10. For each embedding configuration, three runs were performed and the results averaged, resulting in a total of 270 individual runs. We publicly share the code required to reproduce this research as well as all files with calculated results\footnote{\url{https://github.com/dkoterwa/optimizing_bertopic}}. We recommend that users of our artifacts adhere to ethical guidelines and properly cite our work when using these resources.

\section{Results}

\begin{table*}[h]
    \centering
    \scriptsize 
    \setlength{\tabcolsep}{3pt} 
    \renewcommand{\arraystretch}{0.85} 
    \resizebox{0.85\textwidth}{!}{ 
    \begin{tabular}{*{11}{c}}
        \toprule
        & & & \multicolumn{6}{c}{Dataset} \\
        \cmidrule(lr){4-9}
        Method & Configuration & Pooling & \multicolumn{2}{c}{Trump Tweets} & \multicolumn{2}{c}{United Nations} & \multicolumn{2}{c}{20 Newsgroups} \\
        \cmidrule(lr){4-5} \cmidrule(lr){6-7} \cmidrule(lr){8-9}
        & & & TC & TD & TC & TD & TC & TD \\
        \midrule
        \multirow{18}{*}{BERTopic} & Embedding Layer & Max & 0.06 & 0.648 & 0.039 & \textbf{0.886} & 0.104 & 0.856 \\
        & & Mean & 0.063 & 0.695 & \textbf{0.07} & 0.556 & \textbf{0.144} & 0.819 \\
        & & CLS & -0.032 & 0.183 & -0.038 & 0.083 & 0.01 & 0.709  \\
        \cmidrule(lr){2-9}
        & Second Last Layer & Max & 0.07 & 0.676 & 0.047 & 0.488 & 0.118 & 0.861 \\
        & & Mean & 0.066 & 0.65 & 0.039 & 0.456 & 0.109 & 0.829 \\
        & & CLS & 0.039 & 0.553 & 0.016 & 0.473 & 0.056 & 0.703 \\
        \cmidrule(lr){2-9}
        & Last Layer & Max & 0.064 & 0.676 & 0.043 & 0.455 & 0.142 & 0.824 \\
        & & Mean (default) & 0.061 & 0.667 & 0.035 & 0.437 & 0.141 & 0.815 \\
        & & CLS & 0.059 & 0.656 & 0.04 & 0.441 & 0.129 & 0.777 \\
        \cmidrule(lr){2-9}
        & Concat Last Four Layers & Max & \textbf{0.072} & 0.668 & 0.05 & 0.467 & 0.141 & 0.839 \\
        & & Mean & 0.068 & 0.665 & 0.042 & 0.465 & 0.125 & 0.804 \\
        & & CLS & 0.061 & 0.66 & 0.037 & 0.434 & 0.129 & 0.774 \\
        \cmidrule(lr){2-9}
        & Sum Last Four Layers & Max & 0.067 & 0.679 & 0.051 & 0.482 & 0.129 & \textbf{0.942} \\
        & & Mean & 0.069 & 0.666 & 0.042 & 0.465 & 0.12 & 0.82 \\
        & & CLS & 0.062 & 0.66 & 0.036 & 0.433 & 0.129 & 0.778 \\
        \cmidrule(lr){2-9}
        & Sum All Layers & Max & \textbf{0.072} & 0.693 & 0.059 & 0.501 & 0.099 & 0.931 \\
        & & Mean & 0.064 & \textbf{0.696} & 0.052 & 0.619 & 0.125 & 0.858 \\
        & & CLS & 0.061 & 0.652 & 0.033 & 0.429 & 0.125 & 0.77 \\
        \midrule
        LDA & & & -0.011$^\dagger$ & 0.502$^\dagger$ & -0.032 & 0.481 & 0.058$^\dagger$ & 0.749$^\dagger$ \\
        NMF & & & 0.009$^\dagger$ & 0.379$^\dagger$ & 0.002 & 0.305 & 0.089$^\dagger$ & 0.663$^\dagger$ \\
        \bottomrule
    \end{tabular}
    }
    \caption{Topic coherence (TC) and topic diversity (TD) calculated for evaluation datasets. Highest value of topic coherence and topic diversity has been bolded for each dataset. Each score is the average of 15 runs, with 3 runs conducted at topic counts of 10, 20, 30, 40, and 50.  $^{\dagger}$Results come from \citet{grootendorst2022bertopic}.}
    \label{tab:table2}
\end{table*}

Stop word removal is a common preprocessing step in topic modeling, and several studies have examined its impact \citep{schofield-etal-2017-pulling, xu-tackling-topic-general-words}. Our experiments were conducted on datasets both with and without stop words. For stop word removal, we used the NLTK library \citep{bird2006nltk} and its default stop words list. A comparison of performance differences between these dataset versions is presented in \ref{subsec:stopwords}.

\subsection{Differences between configurations}
\label{subsec:differences}
Topic coherence and topic diversity metrics for each dataset, pooling method, and embedding configuration are presented in Table \ref{tab:table2}. Analyzing results across 18 different settings allows us to assess variability and determine whether any configuration outperforms the default one used in BERTopic. Below we present conclusions drawn from Table \ref{tab:table2}:

\begin{enumerate}
  \item Embedding Layer (CLS pooling) performs the worst in terms of topic coherence across all datasets, with negative values for Trump Tweets (-0.032) and United Nations (-0.038).
  \item On the other hand, Embedding Layer (Mean pooling) records the highest value of topic coherence for United Nations (0.07) and 20 Newsgroups (0.144)
  \item CLS pooling is performing the worst in 14 out of 18 experiments (77.8\%) concerning topic coherence and in 16 out of 18 (88.9\%) experiments in terms of topic diversity. An experiment refers to testing a particular embedding configuration and pooling on a specific dataset.
  \item Max pooling guarantees the highest value of topic diversity in 15 out of 18 experiments (83.3\%).
  \item There exists no configuration that yields significantly better outcomes across all datasets.
  \item For all datasets, we are always able to find a configuration that offers higher topic coherence and topic diversity values than the default option used in BERTopic (output from the last layer with Mean pooling).
  \item The default configuration of BERTopic surpasses traditional topic modeling techniques like LDA and NMF across all evaluation metrics on benchmark datasets, except for topic diversity on the United Nations dataset, where LDA scores higher. However, by employing intermediate layer representations, BERTopic not only improves its default performance but also exceeds LDA in achieving greater topic diversity on this specific dataset.
\end{enumerate}

\subsection{Influence of stop words removal}
\label{subsec:stopwords}

\begin{table*}[h]
    \centering
    \small 
    \setlength{\tabcolsep}{3pt} 
    \renewcommand{\arraystretch}{0.85} 
    \resizebox{0.85\textwidth}{!}{ 
    \begin{tabular}{*{11}{c}} 
        \toprule
        & & & \multicolumn{6}{c}{Dataset} \\
        \cmidrule(lr){4-9}
        Method & Configuration & Pooling & \multicolumn{2}{c}{Trump Tweets} & \multicolumn{2}{c}{United Nations} & \multicolumn{2}{c}{20 Newsgroups} \\
        \cmidrule(lr){4-5} \cmidrule(lr){6-7} \cmidrule(lr){8-9}
        & & & TC & TD & TC & TD & TC & TD \\
        \midrule
        \multirow{18}{*}{BERTopic} & Embedding Layer & Max & 0.082 & 0.844 & 0.019 & \textbf{0.873} & 0.116 & 0.893 \\
        & & Mean & 0.046 & 0.848 & 0.161 & 0.807 & 0.163 & 0.861 \\
        & & CLS & -0.165 & 0.673 & -0.008 & 0.224 & 0.014 & 0.694 \\
        \cmidrule{2-9}
        & Second Last Layer & Max & 0.092 & 0.846 & 0.152 & 0.808 & 0.134 & 0.881 \\
        & & Mean & 0.091 & 0.857 & 0.152 & 0.809 & 0.155 & 0.851 \\
        & & CLS & 0.028 & 0.784 & 0.092 & 0.688 & 0.088 & 0.7 \\
        \cmidrule{2-9}
        & Last Layer & Max & 0.078 & 0.875 & 0.148 & 0.763 & 0.169 & 0.851 \\
        & & Mean (default) & 0.067 & 0.863 & 0.144 & 0.765 & 0.164 & 0.842 \\
        & & CLS & 0.069 & 0.858 & 0.138 & 0.754 & 0.152 & 0.824 \\
        \cmidrule{2-9}
        & Concat Last Four Layers & Max & 0.089 & 0.843 & 0.146 & 0.774 & 0.153 & 0.89 \\
        & & Mean & 0.086 & 0.861 & 0.148 & 0.789 & 0.157 & 0.827 \\
        & & CLS & 0.064 & 0.864 & 0.137 & 0.744 & 0.15 & 0.812 \\
        \cmidrule{2-9}
        & Sum Last Four Layers & Max & \textbf{0.095} & 0.864 & 0.154 & 0.786 & \textbf{0.183} & 0.961 \\
        & & Mean & 0.084 & 0.853 & 0.155 & 0.804 & 0.14 & 0.873 \\
        & & CLS & 0.068 & 0.863 & 0.139 & 0.757 & 0.148 & 0.818 \\
        \cmidrule{2-9}
        & Sum All Layers & Max & \textbf{0.095} & 0.868 & 0.154 & 0.781 & 0.092 & \textbf{0.975} \\
        & & Mean & 0.067 & 0.869 & \textbf{0.163} & 0.834 & 0.15 & 0.896 \\
        & & CLS & 0.053 & 0.847 & 0.138 & 0.738 & 0.147 & 0.81 \\
        \midrule
        LDA & & & -0.067 & \textbf{0.925} & 0.045 & 0.758 & 0.063 & 0.75 \\
        NMF & & & 0.025 & 0.58 & 0.083 & 0.597 & 0.089 & 0.666 \\
        \bottomrule
    \end{tabular}
    }
    \caption{Topic coherence (TC) and topic diversity (TD) calculated for evaluation datasets without stop words. Highest value of topic coherence and topic diversity has been bolded for each dataset. Each score is the average of 15 runs, with 3 runs conducted at topic counts of 10, 20, 30, 40, and 50.}
    \label{tab:table3}
\end{table*}

Removing stop words led to a shift in the best-performing configuration across nearly all scenarios for both topic coherence and topic diversity. The results for this experimental setting are shown in Table \ref{tab:table3}. The Embedding Layer configuration no longer achieves the highest topic coherence in any setting. Configurations based on embedding aggregation rank higher after stop words are removed and outperform others in topic coherence for two out of three datasets. The presence of stop word noise appears to hinder the ability of higher transformer layers and aggregation mechanisms to construct clean semantic representations. This limitation is mitigated when stop words are removed, enabling more accurate and diverse topic modeling. The default BERTopic configuration does not emerge as the best in any scenario after stop word removal. There is always an alternative setting that ensures higher topic coherence and topic diversity. 

\begin{figure*}[h!]
    \centering
    \begin{subfigure}{0.48\textwidth}
        \centering
        \includegraphics[width=\linewidth]{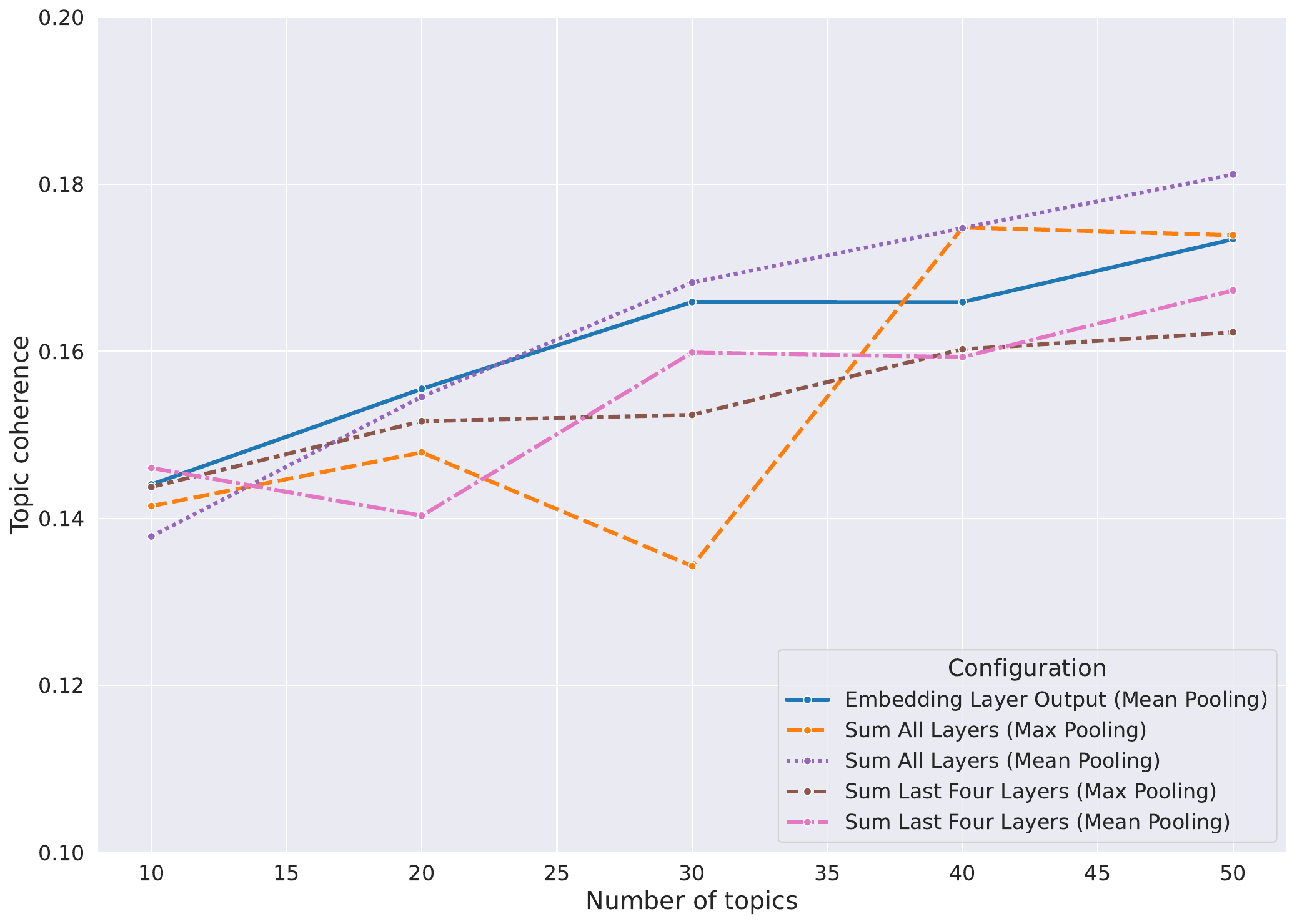}
        \caption{Topic coherence scores.}
    \end{subfigure}
    \hfill
    \begin{subfigure}{0.48\textwidth}
        \centering
        \includegraphics[width=\linewidth]{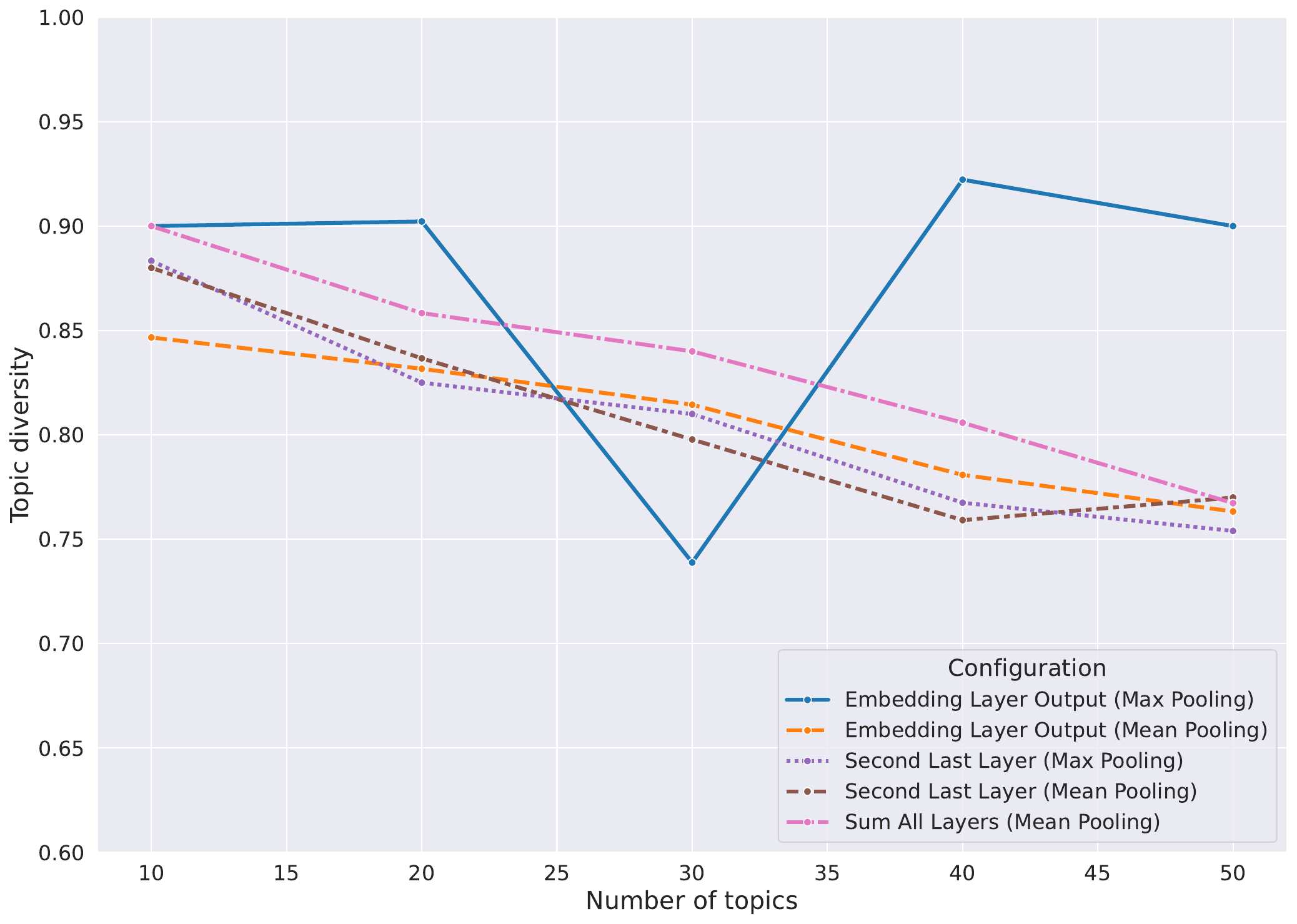}
        \caption{Topic diversity scores.}
    \end{subfigure}
    \caption{Topic coherence and topic diversity scores on the United Nations dataset (without stop words) across number of topics (10–50, step 10) for the 5 best-performing configurations.}
    \label{fig:fig2}
\end{figure*}

The CLS pooling strategy remains the worst performer after stop word removal, achieves the lowest topic coherence in 15 out of 18 experiments (83.3\%) and the lowest topic diversity in 16 out of 18 experiments (88.9\%). Other pooling methods (Mean and Max) aggregate information from the entire input sequence, helping BERTopic produce more coherent and diverse topic representations.
Across a total of 54 experiments, comprising six configuration types, three datasets, and three pooling strategies, stop word removal improves topic coherence in 48 cases (88.9\%) and increases topic diversity in 51 cases (94.4\%). This supports the conclusion that removing stop words enhances BERTopic’s performance. LDA and NMF still fall behind BERTopic, however, LDA achieves the highest topic diversity across all methods on Trump Tweets dataset.
The values in Table \ref{tab:table3} present aggregated scores over different topic counts. Figure \ref{fig:fig2} displays the values of topic coherence and topic diversity across varying numbers of topics on the United Nations dataset (without stop words) for the five best-performing configurations. Increasing the number of topics generally leads to higher topic coherence, while reducing topic diversity, except for the Embedding Layer Output with Max pooling, which does not follow this trend for topic diversity.

\subsection{Visualizing topics from different configurations}
\begin{figure*}[h!]
    \centering
    \begin{subfigure}{0.48\textwidth}
        \centering
        \includegraphics[width=\linewidth]{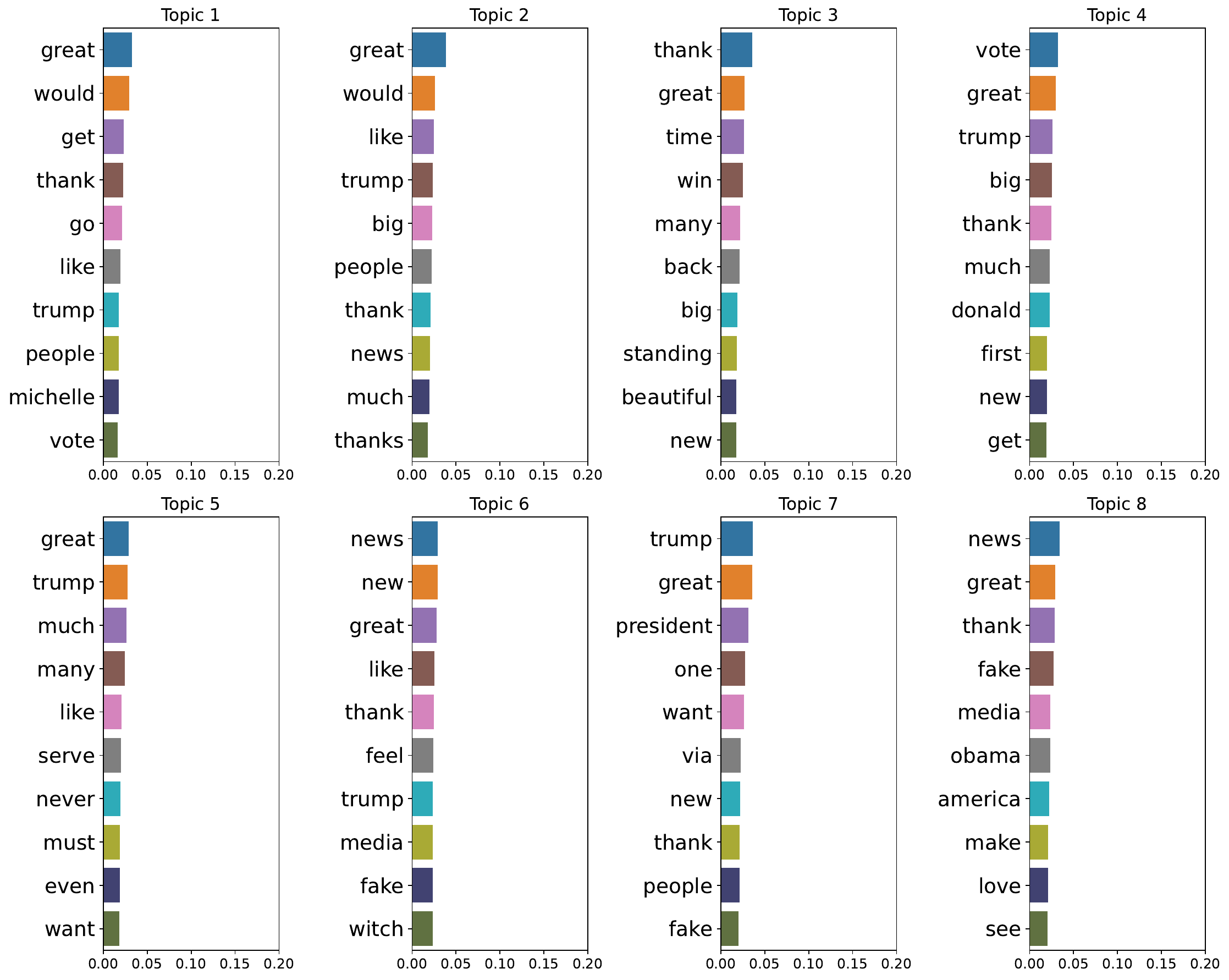}
        \caption{Topic word scores for Embedding Layer with CLS pooling configuration.}
    \end{subfigure}
    \hfill
    \begin{subfigure}{0.48\textwidth}
        \centering
        \includegraphics[width=\linewidth]{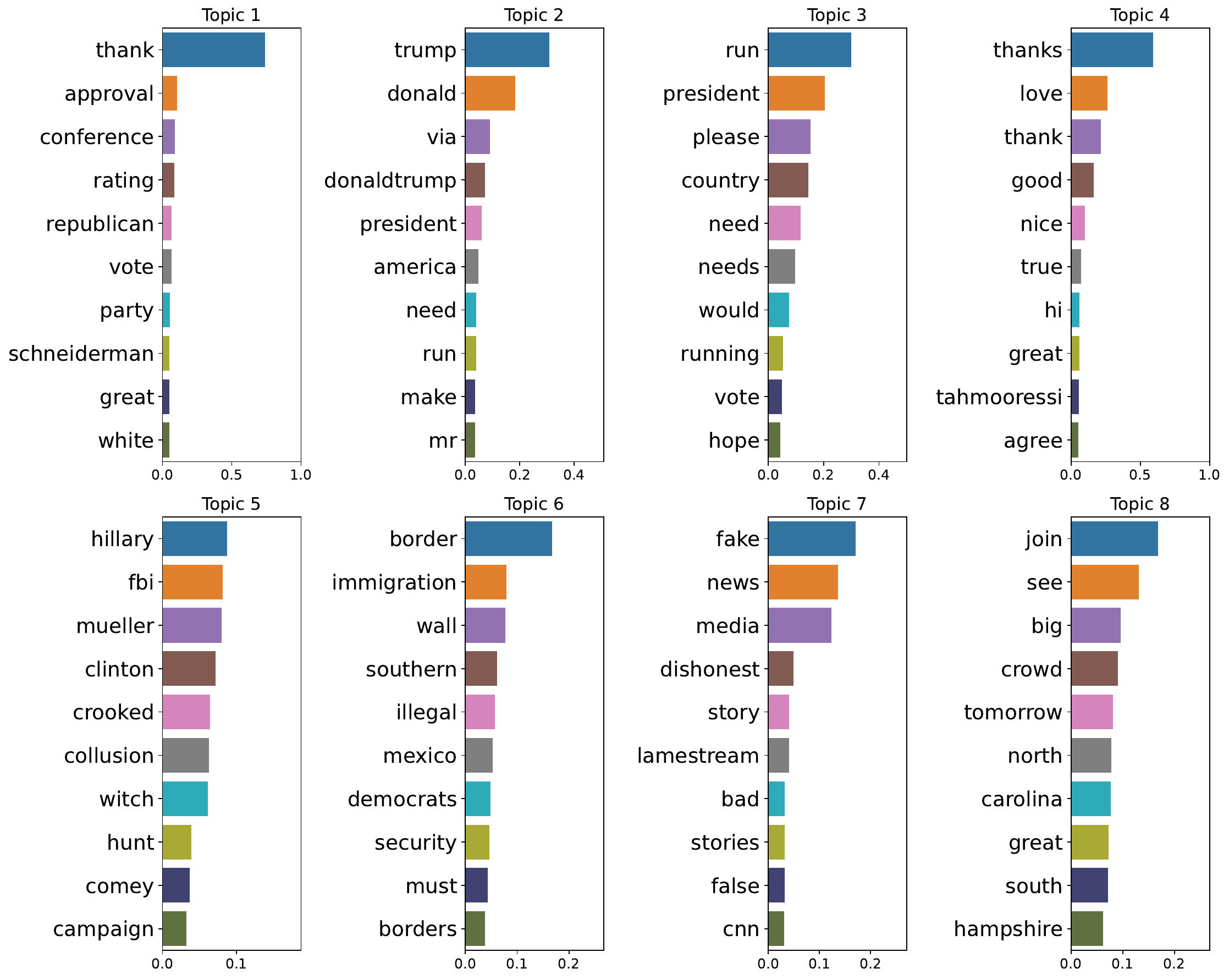}
        \caption{Topic word scores for Sum All Layers with Max pooling configuration.}
    \end{subfigure}
    \caption{Visualization of the top eight topics with words representing them for the worst (Embedding Layer with CLS Pooling) and best (Sum All Layers with Max pooling) configurations.}
    \label{fig:fig3}
\end{figure*}

Visualizing the structure of estimated topics, such as those generated by BERTopic, is essential to gain insights into how the model operates and evaluate its performance. Assessing the quality of topic models based solely on metric values (e.g., coherence or diversity scores) can be challenging, as these metrics may not fully capture the interpretability or relevance of the extracted topics. By visualizing the topics, users can better understand the structure and coherence of the generated themes, which helps in assessing the model's effectiveness.
\newline\indent
For worst and best performing configuration in terms of topic coherence for Trump Tweets without stop words, we present top 8 topics and most influential words for each of them in Figure \ref{fig:fig3}. Difference in the structure of topics is clearly visible. Word scores for worst setting (Embedding Layer with CLS pooling) are more uniformly distributed. It means that there are no sets of words notably characterizing specific topic. This configuration also achieved low value of topic diversity (0.673). Several words, such as "trump", "great", "people" or "thank" repeat in many topics. Based on results from this configuration, we cannot draw valuable conclusions. On the other hand, the best setting (Sum All Layers with Max pooling) provides clear structure of resulting topics. High topic diversity (0.868) assures variability in topics. It is easy to infer from them that Donald Trump mentions immigration on the border with Mexico (Topic 6) or fake news (Topic 7) in his tweets.

\subsection{Dynamic Topic Modeling}
Dynamic topic modeling (DTM) encompasses techniques designed to analyze how topics evolve over time, providing insights into their representation across different time periods. This approach has been widely used in various scientific research contexts \cite{zosa-granroth-wilding-2019-multilingual, zhang-etal-2015-market, sleeman-2021-cybersec}.
\begin{table}[h!]
    \centering
    \footnotesize
    \setlength{\tabcolsep}{6pt} 
    \resizebox{\columnwidth}{!}{ 
    \begin{tabular}{l l l *{4}{c}} 
        \toprule
        & & & \multicolumn{4}{c}{Dataset} \\
        \cmidrule(lr){4-7}
        Method & Configuration & Pooling & \multicolumn{2}{c}{Trump Tweets} & \multicolumn{2}{c}{United Nations} \\
        \cmidrule(lr){4-5} \cmidrule(lr){6-7}
        & & & TC & TD & TC & TD \\
        \midrule
        \multirow{18}{*}{BERTopic} & Embedding Layer & Max & 0.085 & 0.842 & 0.009 & \textbf{0.903} \\
        & & Mean & 0.096 & 0.881 & \textbf{0.174} & 0.856 \\
        & & CLS & -0.099 & \textbf{0.945} & -0.087 & 0.760 \\
        \cmidrule(lr){2-7}
        & Second Last Layer & Max & 0.093 & 0.865 & 0.163 & 0.85 \\
        & & Mean & 0.084 & 0.859 & 0.142 & 0.811 \\
        & & CLS & 0.05 & 0.812 & 0.03 & 0.673 \\
        \cmidrule(lr){2-7}
        & Last Layer & Max & 0.095 & 0.869 & 0.144 & 0.808 \\
        & & Mean (default) & 0.082 & 0.875 & 0.132 & 0.78 \\
        & & CLS & 0.076 & 0.877 & 0.131 & 0.805 \\
        \cmidrule(lr){2-7}
        & Concat Last Four Layers & Max & \textbf{0.102} & 0.846 & 0.152 & 0.815 \\
        & & Mean & 0.078 & 0.863 & 0.139 & 0.798 \\
        & & CLS & 0.073 & 0.863 & 0.122 & 0.796 \\
        \cmidrule(lr){2-7}
        & Sum Last Four Layers & Max & 0.090 & 0.855 & 0.155 & 0.834 \\
        & & Mean & 0.074 & 0.872 & 0.150 & 0.795 \\
        & & CLS & 0.076 & 0.854 & 0.122 & 0.795 \\
        \cmidrule(lr){2-7}
        & Sum All Layers & Max & 0.099 & 0.853 & 0.157 & 0.843 \\
        & & Mean & 0.095 & 0.867 & 0.041 & 0.813 \\
        & & CLS & 0.077 & 0.852 & 0.123 & 0.790 \\
        \midrule
        LDA Sequence & & & 0.09$^\dagger$ & 0.715$^\dagger$ & 0.173$^\dagger$ & 0.82$^\dagger$ \\

        \bottomrule
    \end{tabular}
    }
    \caption{Topic coherence (TC) and topic diversity (TD) for Trump Tweets and United Nations datasets in the dynamic topic modeling setting. Highest value of topic coherence and topic diversity has been bolded for each dataset. Scores were computed at each of the 9 timesteps per dataset, and then averaged over 3 runs for every timestep. $^{\dagger}$Results come from \citet{grootendorst2022bertopic}.}
    \label{tab:table4}
\end{table}

\begin{figure*}[h!]
    \centering
    \includegraphics[scale=0.5]{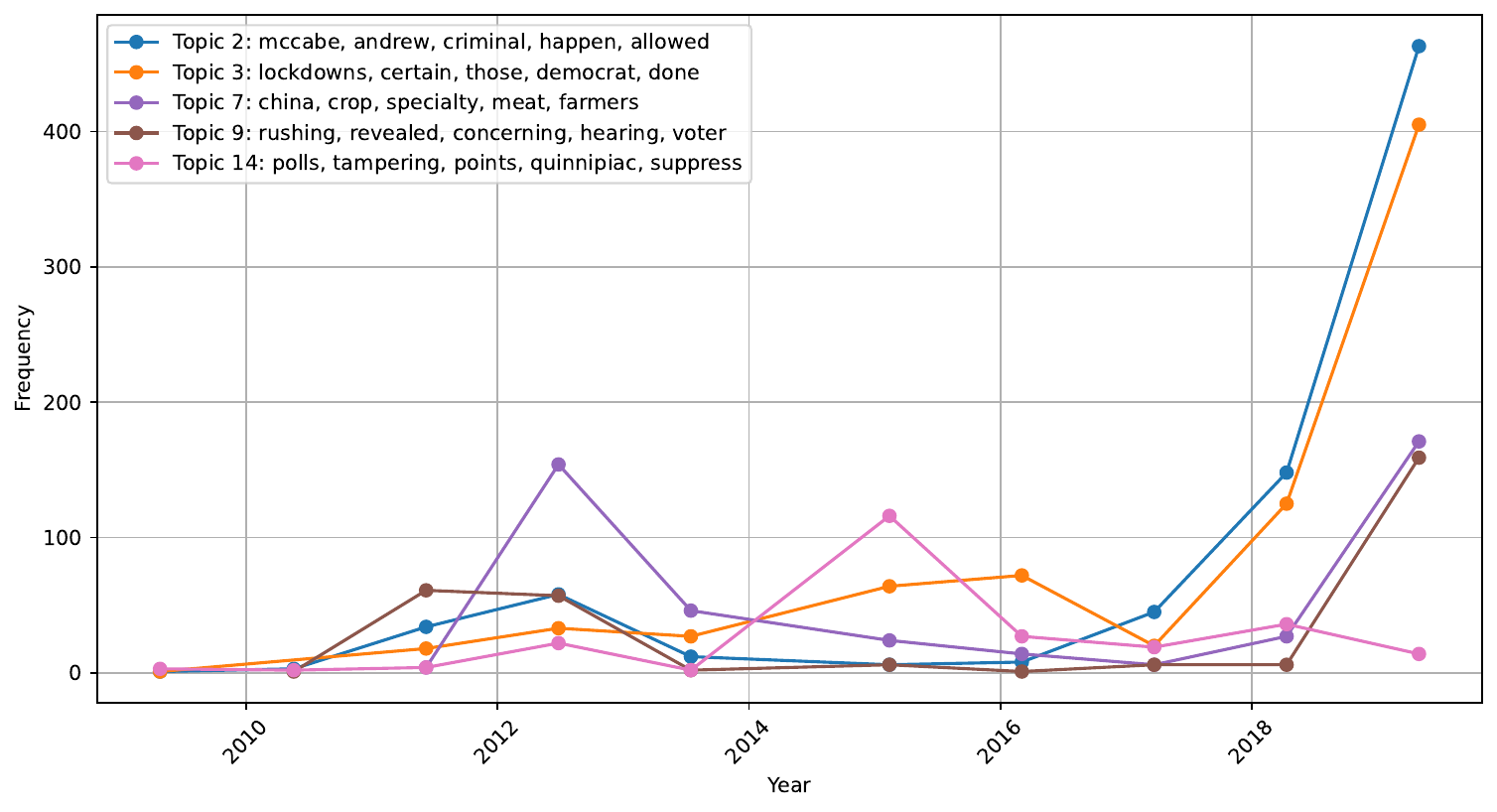}
    \caption{Frequency of selected topics estimated from Trump Tweets dataset over the years. Topics have been created by utilizing the best configuration in terms of topic coherence (Concat Last Four Layers with Max pooling).}
    \label{fig:fig4}
\end{figure*}

We evaluated all configurations in the dynamic topic modeling scenario using the Trump Tweets and United Nations datasets and present the results in Table \ref{tab:table4}. Most findings align with those in \ref{subsec:differences}. CLS pooling consistently performs the worst, while Max pooling achieves the highest performance in most experiments. For each dataset, there is always a configuration that outperforms the default BERTopic setting. The traditional dynamic topic modeling approach (LDA Seq) achieves better topic coherence on two datasets and higher topic diversity on one compared to the default BERTopic configuration. However, using intermediate layer representations in BERTopic consistently results in higher metric values across all evaluated datasets and outperforms the traditional algorithm.

\indent
Dynamic topic modeling, facilitated by BERTopic, enables the analysis of topic evolution over time. Figure \ref{fig:fig4} illustrates the frequency of selected topics from the Trump Tweets dataset over different years, providing information on their temporal representation.

\section{Conclusion}
\label{conclusions}
In this work, we explore topic modeling with BERTopic by leveraging representations from different layers of the Transformer model. We demonstrate that searching across various embedding configurations enhances both topic coherence and diversity in traditional and dynamic topic modeling. We analyze the impact of stop words removal and conclude that eliminating stop words from the text corpus improves metric values in nearly all cases. Additionally, we show that the choice of pooling method plays a crucial role in the final topic modeling results, with significant differences in topic structures between the best and worst performing configurations. In future work, we aim to investigate how BERTopic utilizes knowledge from specific layers to better understand its role in shaping topic structures.

\section{Limitations}
Searching through embedding configurations leads to better results but comes at the cost of increased evaluation time, which is a limitation of this approach. For instance, evaluating all 18 configurations on the Trump Tweets dataset takes approximately 1.5 hours. A key limitation of topic modeling is the need for manual assessment of the estimated topics, which makes the evaluation time consuming. This issue is further compounded when comparing multiple configurations and analyzing differences in topic structures. This limitation could be mitigated in future work by employing a Large Language Model to automatically assess the quality of the estimated topics--thereby reducing the reliance on manual evaluation.
\newline \indent 
Another limitation is the lack of interpretability. Although many configurations outperform the default setting in terms of metrics, understanding why and how they achieve better results remains a challenge. As mentioned in \ref{conclusions}, we aim to explore this aspect in future work.

\bibliography{acl_latex}



\appendix
\section{Computational Resources and Model Details}

\subsection{Model Parameters}
In our experiments, we used the following models together with their parameter counts:
\begin{itemize}
    \item \textbf{all-mpnet-base-v2}: 110 million parameters
\end{itemize}

\subsection{Computational Budget}
The total computational budget for our experiments is summarized as follows:
\begin{itemize}
    \item \textbf{Total GPU hours used}: 4 hours
    \item \textbf{Total CPU hours used}: 168 hours
\end{itemize}

\subsection{Computing Infrastructure}
Our experiments were carried out using a Nvidia L4 GPU with 24GB of virtual memory.
\end{document}